\documentclass[10pt,logo,copyright]{nvidiatechreport}

\ifXeTeX
  \microtypesetup{tracking=false}
  \usepackage{newunicodechar}
  \newunicodechar{–}{\textendash}
  \newunicodechar{—}{\textemdash}
  \newunicodechar{’}{\textquoteright}
  \newunicodechar{×}{\ensuremath{\times}}
  \newunicodechar{≈}{\ensuremath{\approx}}
  \newunicodechar{≤}{\ensuremath{\leq}}
  \newunicodechar{≥}{\ensuremath{\geq}}
  \newunicodechar{−}{\ensuremath{-}}
  \newunicodechar{č}{\v{c}}
\fi

\usepackage{amsmath}
\usepackage{amssymb}
\usepackage{booktabs}
\usepackage[round]{natbib}
\usepackage{xspace}
\usepackage{multirow}
\usepackage{makecell}
\usepackage{tabularx}
\usepackage{longtable}
\usepackage{bm}
\usepackage{listings}
\usepackage{titletoc}
\usepackage{placeins}
\usepackage{needspace}
\usepackage[capitalise,noabbrev]{cleveref}

\definecolor{nvgreen}{HTML}{76B900}
\hypersetup{allcolors=nvgreen, citecolor=black,
            pdftitle={NVIDIA Kumo Tabular: Pushing the Accuracy--Efficiency Frontier for Tabular Prediction},
            pdfauthor={Kumo Tabular Team}}
\newcommand{\modelname}{\mbox{Kumo~Tabular}\xspace}
\DeclareRobustCommand{\kumoS}{\mbox{Kumo~Tabular-S}\xspace}
\DeclareRobustCommand{\kumoM}{\mbox{Kumo~Tabular-M}\xspace}
\DeclareRobustCommand{\kumoL}{\mbox{Kumo~Tabular-L}\xspace}
\newcommand{\sdm}{SDM\xspace}
\newcommand{\priorname}{Kumo prior\xspace}

\newcommand{\includefig}[1]{\IfFileExists{#1}{\includegraphics{#1}}{\fbox{\parbox{0.9\linewidth}{\centering\vspace{1cm}Missing figure: \texttt{\detokenize{#1}}\vspace{1cm}}}}}

\definecolor{kumogreen}{HTML}{00785A}

\definecolor{codebackground}{HTML}{F7F8FA}
\definecolor{codekeyword}{HTML}{0072B2}
\definecolor{codestring}{HTML}{00785A}
\definecolor{codecomment}{HTML}{7F7F7F}
\lstdefinestyle{sdm}{
  language=Python,
  basicstyle=\small\ttfamily,
  keywordstyle=\color{codekeyword},
  stringstyle=\color{codestring},
  commentstyle=\color{codecomment}\itshape,
  backgroundcolor=\color{codebackground},
  numbers=none,
  xleftmargin=1.5em,
  framexleftmargin=1em,
  frame=single,
  framerule=0pt,
  framesep=6pt,
  showstringspaces=false,
  columns=fullflexible,
  keepspaces=true,
  breaklines=true,
  captionpos=t
}

\newcommand{\taLimixPredictRatio}{36}

\newcommand{\baTabpfnPredictRatio}{1.7}

\title{NVIDIA Kumo Tabular: Pushing the Accuracy--Efficiency Frontier for Tabular Prediction}

\usepackage{authblk}

\author[1]{Jingang Qu}
\author[1]{Valter Hudovernik}
\author[1]{Martin Jurkovič}
\author[1]{Cedric Lorenz}
\author[1,2]{David Holzmüller}
\author[1]{Federico Lopez}
\author[1]{Dmitry Gordeev}
\author[1]{Akihiro Nitta}
\author[1]{Ramona Bendias}
\author[1]{Ardrian Wong}
\author[1]{Vedaant Jain}
\author[1,3]{Vignesh Kothapalli}
\author[1]{Gilberto Titericz Junior}
\author[1,3]{Jure Leskovec}
\author[1]{Matthias Fey}
\affil[1]{NVIDIA}
\affil[2]{INRIA}
\affil[3]{Stanford University}

\begin{abstract}
\textbf{Abstract.}\par
We introduce \modelname, a family of open tabular foundation models for classification and regression, in three sizes: \kumoS (27M parameters), \kumoM (61M), and \kumoL (214M).
Pretrained only on artificial tables from the \priorname, they can make predictions on new tables without any training.
In our approach, a table encoder compresses each row into an embedding, and an in-context learning transformer predicts the unlabeled rows from the labeled ones.
\modelname introduces architectural innovations that improve both training stability and predictive performance, enabling the model to scale up to 150M cells on modern GPUs.
\kumoL ranks first on eight benchmarks: TabArena, TALENT, BeyondArena, ScoringBench, STRABLE, MulTaBench, RamanBench, and \mbox{TabBench-Bio}.
These cover probabilistic regression, tables with over 1M rows or 100K columns, and tables with strings, text, or images.
Accuracy improves with model size in all tested comparisons.
All three sizes lie on the accuracy--speed frontiers of TabArena and BeyondArena.
\kumoL predicts faster than the runners-up, \mbox{LimiX-2} on TabArena (\taLimixPredictRatio$\times$) and \mbox{TabPFN-3.5} on BeyondArena (\baTabpfnPredictRatio$\times$).
Weights and pretraining code are released under OpenMDW-1.1 and Apache 2.0, respectively.

\end{abstract}

\begin{document}
\maketitle

\suppressfloats[t]
\par\vspace{6pt}
{\small\noindent
\begin{tabularx}{\linewidth}{@{}X l r@{}}
\toprule
\textbf{Code:} Pretraining, inference, \priorname & \href{https://github.com/NVIDIA/structured-data-models}{\texttt{github.com/NVIDIA/structured-data-models}} & Apache-2.0 \\
\textbf{Weights:} Classifier and regressor in all sizes & \href{https://huggingface.co/nvidia/Kumo-Tabular}{\texttt{huggingface.co/nvidia/Kumo-Tabular}} & OpenMDW-1.1 \\
\bottomrule
\end{tabularx}\par}

\section{Introduction}
\label{sec:introduction}

\begin{figure}[t]
  \centering
  \includefig{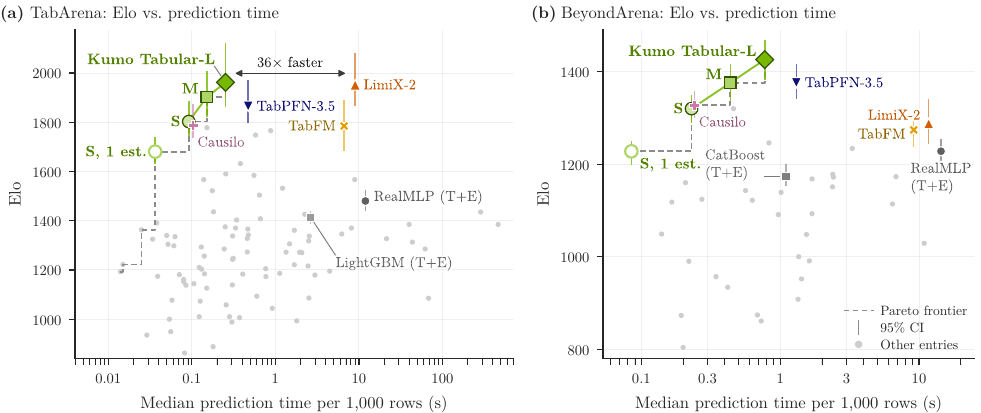}
  \caption{\textbf{\modelname is accurate and fast.}
    On (a)~TabArena and (b)~BeyondArena, all three sizes lie on the Pareto frontier of Elo and prediction time.
    }
  \label{fig:speed}
\end{figure}

Tables are among the most common forms of data in industry and science, from bank transactions, credit histories, and online orders to factory sensor logs, patient records, census data, and laboratory measurements~\citep{benjelloun2020google,shwartzziv2022tabular,borisov2024deep,vanbreugel2024position}.
For two decades, the strongest predictors on such data were gradient-boosted decision trees~\citep{friedman2001greedy,chen2016xgboost,ke2017lightgbm,prokhorenkova2018catboost}, tuned for each task and often ensembled by AutoML systems~\citep{erickson2020autogluon}.
\emph{Tabular foundation models (TFMs)} have changed this picture.
They can make predictions on a new table without training or tuning on it.
Pretrained on a large number of predictive tasks, they read the labeled train rows of the table as context and predict the labels of its test rows in one forward pass~\citep{muller2022transformers,hollmann2023tabpfn}.
The strongest TFMs now rank well ahead of tuned and ensembled gradient-boosted trees and deep tabular networks on TabArena~\citep{erickson2025tabarena} and BeyondArena~\citep{purucker2026beyondiid}, two broad benchmarks of real-world tabular datasets (\cref{fig:speed}).

TabPFN established the TFM paradigm but handled only relatively small classification tables~\citep{hollmann2023tabpfn}.
Later models extended it to regression and to tables with hundreds of thousands of rows~\citep{hollmann2025tabpfn2,qu2025tabicl}, learned from richer artificial priors and from real tables~\citep{ma2026tabdpt,zhang2026mitra,qu2026tabiclv2}, and began to exploit column names with semantics-aware designs~\citep{spinaci2025contexttab}.
TabICLv2 showed that a fully open model can reach the state of the art~\citep{qu2026tabiclv2}, launching a new wave of innovation.
Several models have since pushed accuracy on standard i.i.d.\ benchmarks even further~\citep{grinsztajn2026tabpfn,kong2026tabfm,eo2026exaone,tao2026mitrav2,cho2026causilo,limix2026limix2,jaeger2026tabpfn35}.

Here, we present \emph{\modelname}, a family of fully open TFMs for classification and regression that combine frontier performance with computational efficiency.
It comes in three sizes, \emph{\kumoS}, \emph{\kumoM}, and \emph{\kumoL} (27M, 61M, and 214M parameters), each as a classifier and a regressor.
Its architecture extends TabICL~\citep{qu2025tabicl} and TabICLv2~\citep{qu2026tabiclv2} with innovations that improve both training stability and predictive performance (\emph{cf.}~\cref{sec:model}).
The pretraining data is provided by the \emph{\priorname}, a generator of artificial tables that builds on the prior of TabICLv2 (\emph{cf.}~\cref{sec:prior}).
For each table, the prior draws a random causal graph, the functions that link its nodes, and how the columns are observed from them. The prior also includes realistic noise patterns and missing values.
Pretraining runs in three stages, from tables of 1,024 rows to tables of up to 60K rows, all with up to 100 columns (\emph{cf.}~\cref{sec:training}). We release both the pretraining and inference code in our open-source library \texttt{structured-data-models} (\sdm)\footnote{GitHub: \url{https://github.com/NVIDIA/structured-data-models}}.

\begin{figure}[t]
  \centering
  \includefig{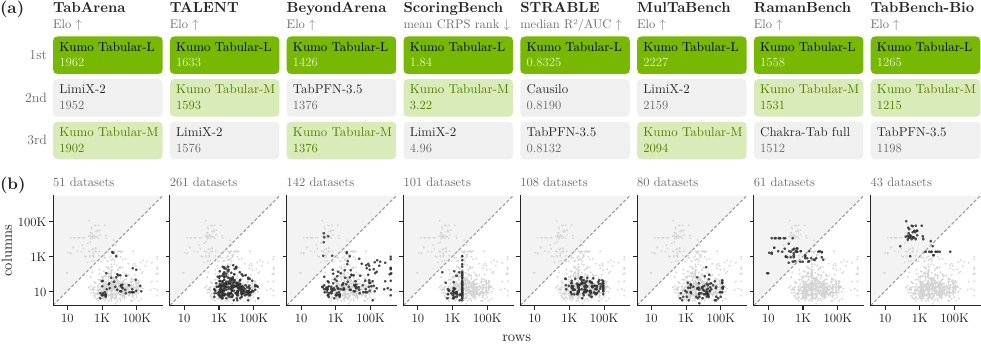}
  \caption{\textbf{\kumoL tops all eight leaderboards.}
    \textbf{(a)}~Entries and scores under each leaderboard's main metric, \modelname in green.
    \textbf{(b)}~Row and column counts of each dataset (log scale). ScoringBench subsamples to 3,000 rows, and STRABLE and MulTaBench count columns before encoding. Tables of each benchmark are colored black over all tables in gray. Above the dashed line, tables have more columns than rows.
    See \cref{tab:app-benchmarks} for benchmark versions, evaluation protocols, ranking procedures, and subsampling limits.
    }
  \label{fig:hero}
\end{figure}

\kumoL ranks first on eight benchmarks, while the strongest entry other than \modelname varies by benchmark (\emph{cf.}~\cref{fig:hero}).
They span broad suites of real tables~\citep[TALENT, TabArena, and BeyondArena;][]{ye2024closer,erickson2025tabarena,purucker2026beyondiid}, probabilistic regression~\citep[ScoringBench;][]{landsgesell2026scoringbench}, tables with strings~\citep[STRABLE;][]{blayer2026strable}, text or images~\citep[MulTaBench;][]{multabench2026leaderboard}, and wide scientific tables~\citep[RamanBench and \mbox{TabBench-Bio};][]{koddenbrock2026ramanbench,kreuer2026tabbenchbio}.
The benchmark datasets reach more than 1M rows and more than 100K columns. 
On an H100 GPU our models may attend to large tables effectively, scaling up to 280M cells for \kumoS and 150M for the larger variants.
On TabArena and BeyondArena, all three sizes lie on the accuracy--speed frontiers (\emph{cf.}~\cref{fig:speed}).
\kumoL also predicts faster than the runner-up on each: \mbox{LimiX-2} (\taLimixPredictRatio$\times$) and \mbox{TabPFN-3.5} (\baTabpfnPredictRatio$\times$).

\section{Model Architecture}
\label{sec:model}
\label{sec:model-overview}
\suppressfloats[t]

\begin{figure}[!tp]
  \centering
  \includegraphics[width=\linewidth]{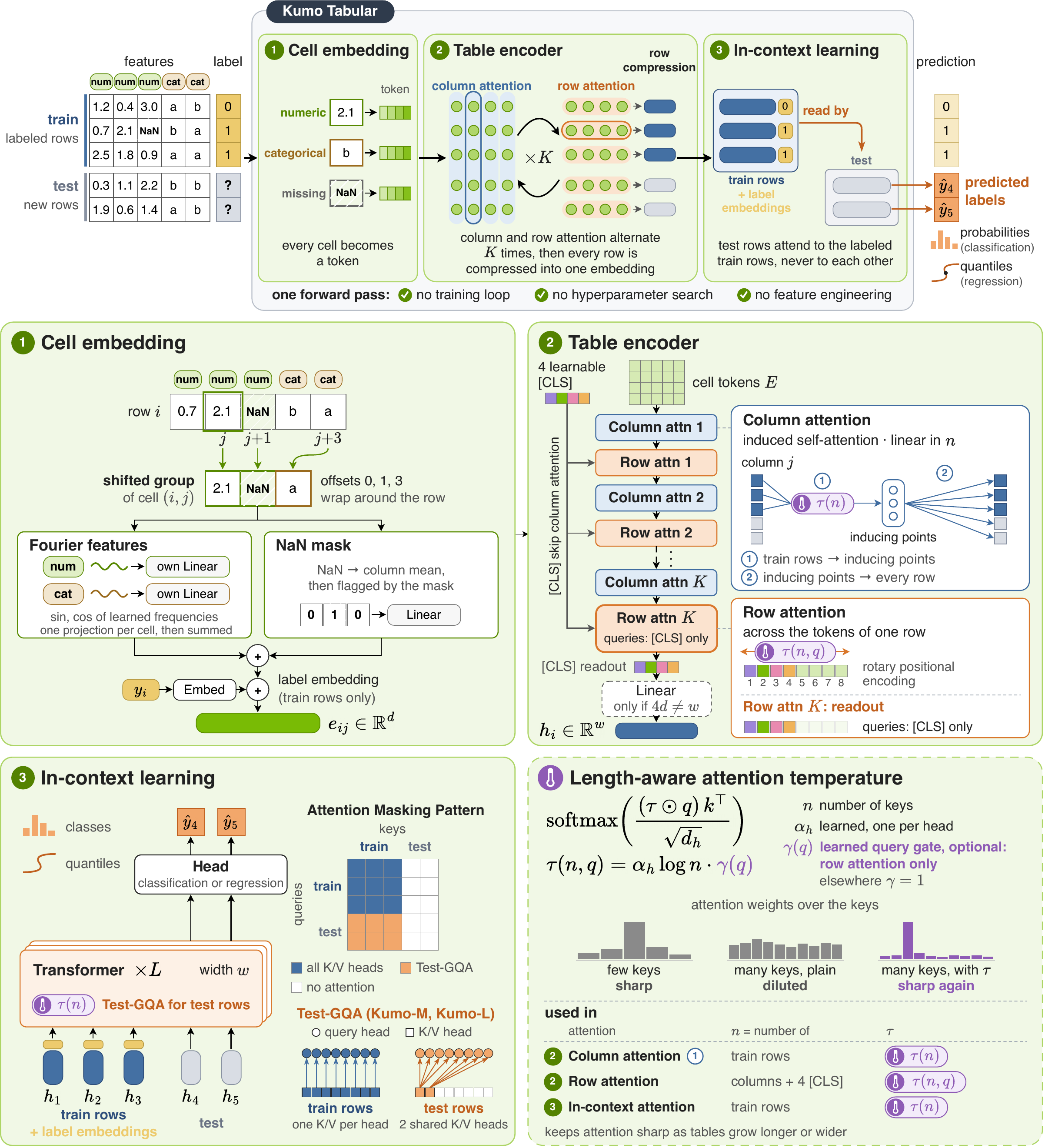}
  \caption{\textbf{Architecture of \modelname.}
  \textbf{(1)}~The token of cell $(i, j)$ encodes the cells in columns $j$, $j{+}1$, and $j{+}3$ of row $i$, their missingness flags, and, for train rows, the label.
  \textbf{(2)}~Column and row attention alternate $K$ times.
  Column attention passes through inducing points that read the train cells of each column. Four readout tokens (\texttt{[CLS]}) form the row embedding, which enters the ICL transformer as $h_i$.
  \textbf{(3)}~Test rows attend only to train rows. In \kumoM and \kumoL, they read two shared key/value heads (Test-GQA).
  Length-aware temperature keeps attention sharp as tables grow longer or wider.
  }
  \label{fig:architecture}
\end{figure}

\modelname predicts the labels of the \emph{unlabeled test rows} of a table from its labeled \emph{training rows} in a single forward pass. %
It processes a table at two granularities: a table encoder builds a fixed-size
embedding for every row by attending within columns and rows of a table, and an \emph{in-context learning (ICL)}
transformer then reasons across rows, allowing test rows to learn from labeled training examples (\emph{cf.}~\cref{fig:architecture}):

\paragraph{Cell Embedding.}
Each cell is represented by a token together with two neighboring cells in its row, following the feature grouping design of TabICLv2~\citep{qu2026tabiclv2}.
As in \citet{kong2026tabfm}, values are expanded into sine and cosine features at learned frequencies and projected based on column type (numerical \emph{vs.} categorical) into $d$-dimensional embeddings.
Missing values are represented by learned embeddings added to their imputed numerical representations.
For training rows, the token also holds an embedding of the row's label.

\paragraph{Table Encoder.}
In most two-stage tabular in-context learning architectures,
the encoder first runs several column attention layers (attention between cells in the same column), followed by several row attention layers (attention between cells in the same row), which prevents the column attention layers from accessing information across columns.
Instead, we \emph{alternate} column and row attention four times in \kumoS and six times in \kumoM and \kumoL. %
Interleaving has a similar runtime to staged column and row attention, but it makes memory optimizations such as row chunking more expensive \citep{grinsztajn2026tabpfn}.
Column attention runs through learned \emph{inducing points}~\citep{lee2019set}, so its cost grows linearly with the number of rows.
These points attend only to the column's training cells, while every cell attends to the inducing points.
Row attention runs over the cell tokens of a row, which it tells apart by rotary position embeddings~\citep{su2024roformer}.
Four learned \emph{readout tokens} join each row in row attention but skip column attention. 
They serve as readout to obtain the final fixed-size row embeddings.

\paragraph{ICL Transformer.} A separate transformer processes the obtained row embeddings. Label embeddings are added to the training row embeddings before the first layer.
Train rows attend only to one another, so their keys and values can be computed once and cached.
To shrink this cache, we use Test-GQA in \kumoM and \kumoL. Test rows thus only read the first two key/value heads of each layer, making the cache 4-fold smaller per train row in \kumoM and 8-fold smaller in \kumoL. %
After the final layer, a two-layer MLP maps each test row to ten class logits or to 999 quantiles of the target.
At inference, 
tasks with more than ten classes are split into ten-class subtasks. %

\paragraph{Model Development.}
We build on the open design of TabICLv2~\citep{qu2026tabiclv2}.
We progressively modify the architecture to improve training stability, efficiency, and predictive performance, then scale up the model.
TabICLv2 introduced \emph{query-aware scalable softmax (QASSMax)}, which scales logits with the logarithm of the number of keys to counter attention dilution as tables grow.
We revised attention normalization and scaling after observing instabilities in longer training runs and larger models.
We also apply length-dependent scaling to row attention to retain sharp attention on wider tables.
We interleave column and row attention to allow interactions across columns throughout the encoder, refine the transformer blocks, and add missing-value-aware embeddings to support our expanded prior.
With the resulting efficiency gains, we increase the encoder capacity and scale the ICL transformer to obtain the three model sizes below.
We evaluate this progression through an additive ablation study in \Cref{sec:ablations} and report our development observations in \cref{app:stability}.

\paragraph{Model Family.}

\begin{table}[t]
  \centering
  \caption{\textbf{The \modelname family.}
  Model configurations of \modelname across model sizes and task types.
  }
  \label{tab:model-family}
  \small
  \begin{tabular}{@{}llccc@{}}
    \toprule
    & & \textbf{\kumoS} & \textbf{\kumoM} & \textbf{\kumoL} \\
    \midrule
    \multicolumn{2}{@{}l}{Parameters, classifier / regressor} & 27.46M / 28.47M & 61.49M / 62.49M & 213.67M / 215.68M \\
    \multicolumn{2}{@{}l}{ICL share of parameters, classifier / regressor} & 93.8\% / 94.1\% & 82.9\% / 83.2\% & 95.3\% / 95.4\% \\
    \midrule
    \multirow{5}{*}{\makecell[l]{Table\\Encoder}}
    & Stages (column block $\rightarrow$ row block) & 4 & 6 & 6 \\
    & Cell width $d$ & 128 & 256 & 256 \\
    & Heads $\times$ head width & $4 \times 32$ & $4 \times 64$ & $4 \times 64$ \\
    & Inducing points $M$ & 128 & 256 & 256 \\
    & Row embedding $4d \rightarrow w$ & 512 & 1{,}024 $\rightarrow$ 512 & 1{,}024 \\
    \midrule
    \multirow{4}{*}{\makecell[l]{ICL\\Transformer}}
    & Layers & 12 & 24 & 24 \\
    & Width $w$ & 512 & 512 & 1{,}024 \\
    & Heads $\times$ head width & $8 \times 64$ & $8 \times 64$ & $16 \times 64$ \\
    & Key/value heads read by test rows & all 8 & 2 of 8 & 2 of 16 \\
    \bottomrule
  \end{tabular}
\end{table}

\modelname comes in three sizes, each with a classifier and a regressor (\emph{cf.}~\cref{tab:model-family}).
\kumoS closely matches TabICLv2 in parameter count and ICL transformer shape.
\kumoM has a wider and deeper table encoder and twice as many ICL layers, and \kumoL doubles the width and heads of the ICL transformer of \kumoM.
Around 83\% to 95\% of the parameters sit in the ICL transformer.

\section{Pretraining Data}
\label{sec:prior}

\begin{figure}[tp]
  \centering
  \includegraphics[width=\linewidth]{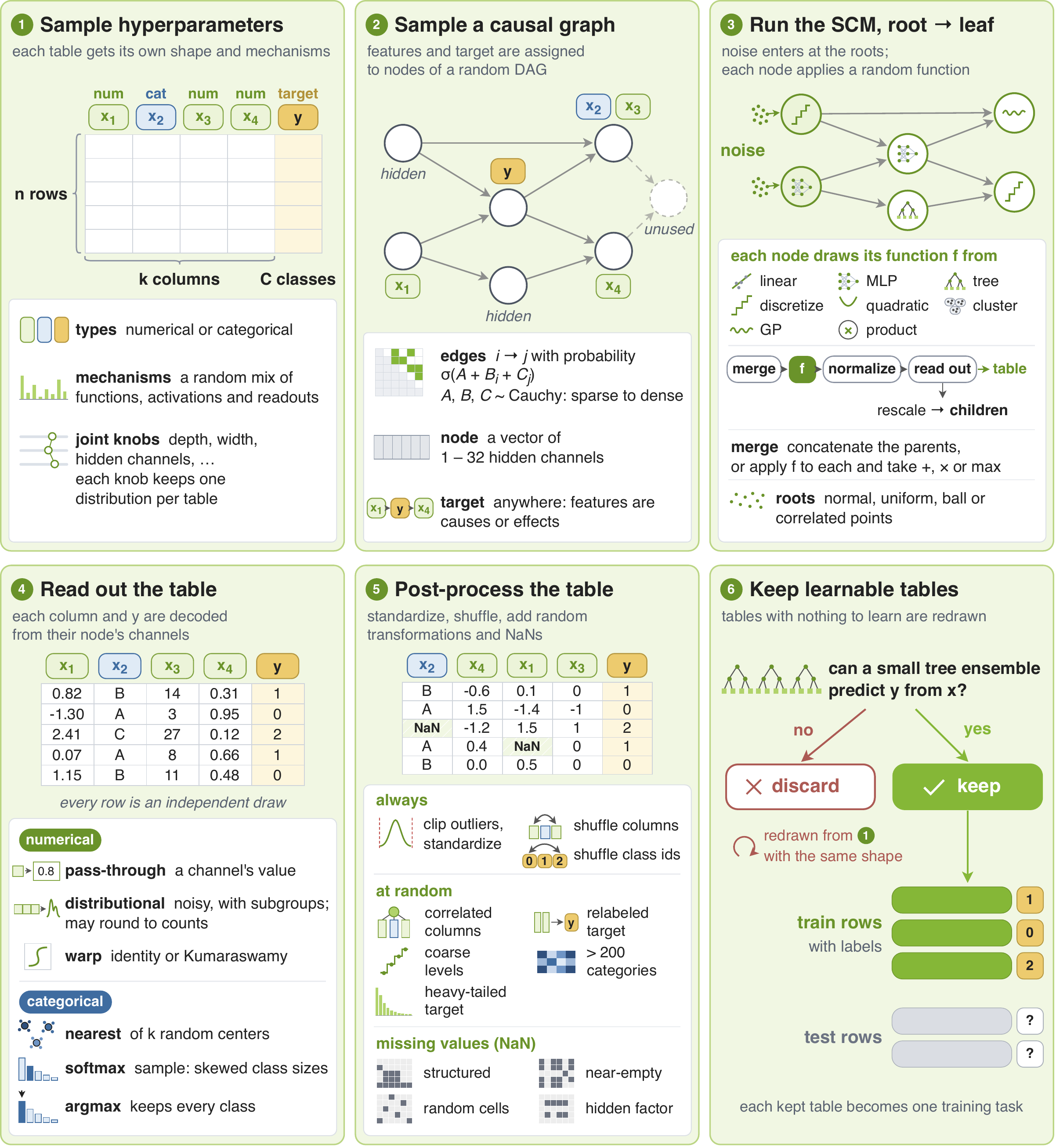}
  \caption{\textbf{The \priorname combines causal dependencies with measurement noise and missing values.}
    Building on TabICLv2~\citep{qu2026tabiclv2}, steps~1--3 sample the table dimensions, variable types, causal graph, and node functions.
    Step~4 maps the resulting node representations to numerical or categorical features and targets.
    Our extensions to the readout and post-processing stages introduce noisy measurements, correlated features, varied feature and target distributions, and missingness patterns (steps~4--5).
    A small tree ensemble filters tables by target predictability (step~6).
    Retained tables are split into labeled train rows and test rows, whose targets supervise pretraining.
    \Cref{app:prior} specifies the new and modified components.}
  \label{fig:prior}
\end{figure}

A TFM trained on tables drawn from a prior learns to approximate the posterior predictive distribution under that prior~\citep{muller2022transformers}, so the prior must produce both the varied dependencies and the imperfections of real tables.
The \priorname builds on the causal generator of TabICLv2~\citep{qu2026tabiclv2}: each table comes from its own random \emph{structural causal model} (SCM;~\citealp{pearl2009causality}) and is generated in six steps (\emph{cf.}~\cref{fig:prior}).
 It adds mechanisms that model properties of real tables that random causal models alone do not reliably reproduce: %
noisy measurements, groups of correlated features, coarse or code-like columns, targets derived from a subset of features, skewed targets, and missing values.
\cref{app:prior} provides a detailed specification of our changes and newly added components.

\paragraph{1. Hyperparameters.}
Each table draws its own hyperparameters: its shape, column types, and individual parameters for subsequent generation steps.
Within a table, these draws follow per-SCM preferences, so its nodes tend to use related mechanisms and hyperparameter settings. 

\paragraph{2. Causal Graph.}
Features and target are assigned to random nodes of a random directed acyclic graph, sparse in some tables and dense in others. %
A feature can thus be a cause or an effect of the target, and nodes without a column stay hidden.

\paragraph{3. Propagation.}
Values flow through the graph in topological order: roots initialize random points, and every other node combines its parents' outputs using random functions and a merging method.
The functions come from eight families, such as small neural networks, tree ensembles, and Gaussian processes. %

\paragraph{4. Readout.}
Each node holds a vector of hidden channels, and converters read each column and the target out of the channels of their corresponding node. Converters can also update these channels with the generated column or target values before passing them to child nodes.
A new \emph{distributional numerical converter} samples numerical features and regression targets from distributions whose parameters depend on the node channels. Three channels control the location, noise scale, and probability of belonging to a shifted subgroup. The sampled values can then be transformed or rounded to counts. %
This produces numerical columns representing noisy measurements or counts, with noise levels and subgroup probabilities that vary across rows.
For categorical features and classification targets, the converter maps node channels to a score for each category, then either samples from the resulting softmax distribution or deterministically selects the highest-scoring category.

\paragraph{5. Post-Processing.}
As in TabICLv2, outliers are clipped, columns are standardized, class codes are permuted, and rows and columns are shuffled.
Three new kinds of transformations apply at random:
\begin{itemize}[leftmargin=1.2em,itemsep=2pt,topsep=3pt]
  \item \textbf{Correlated feature blocks:}
  Real tables often hold groups of related columns, such as the items of a questionnaire.
  In 30\% of the tables, we introduce correlations within randomly selected blocks of feature columns. For each block, we draw one shared random value per row and blend it into the numerical columns~\citep{synthefy2026nori}. For categorical columns, we use this shared value together with the existing categories to reorder category assignments across rows, preserving each category’s frequency.
  \item \textbf{Realism augmentations:}
  Different augmentations can modify a fraction of generated tables: 
  target relabeling makes targets depend on only a few features; feature coarsening allows identical feature rows to have different targets; high-cardinality codes introduce more than 200 categories to mimic identifiers; and, from stage~3 onward, heavy-tailed regression targets mimic quantities such as prices.
  \item \textbf{Missing values:}
  Missing values can reflect how data are collected. A failed source leaves whole rows incomplete, a lab value may be recorded only for patients with a particular diagnosis, and an instrument may not report values outside its measurement range.
  In about a quarter of generated tables, we introduce missing values through one or more patterns~\citep{synthefy2026nori}: missingness across rows, dependence on other features, values outside specified ranges, nearly empty columns, blocks of missing values governed by a shared hidden factor for each row, or independent random missingness.
\end{itemize}

\paragraph{6. Learnability Filter.}
A table is redrawn with the same shape if a small ensemble of extremely randomized trees~\citep{geurts2006extremely} cannot reliably beat predicting the mean.
Additionally, the filter flags \emph{too-easy} tables~\citep{pfeiffer2026tabh2o}, which a tree ensemble predicts with at most 5\% of the error of predicting the mean.

\section{Pretraining}
\label{sec:training}
\suppressfloats[t]

We train each %
checkpoint from scratch on fresh draws from the \priorname in three stages, following TabICLv2 \citep{qu2026tabiclv2}. 
Each stage teaches the model to use longer, more challenging tables (\emph{cf.}~\cref{tab:training}):

\begin{table}[tb]
  \centering
  \caption{\textbf{Training recipe by stage}, for the classifier and the regressor of every model size.
  \kumoS, \kumoM, and \kumoL use 8, 32, and 64 NVIDIA H100 80\,GB GPUs, respectively, in all stages.}
  \label{tab:training}
  \small
  \begin{tabularx}{\linewidth}{@{}>{\raggedright\arraybackslash}p{0.25\linewidth}>{\raggedright\arraybackslash}X>{\raggedright\arraybackslash}X>{\raggedright\arraybackslash}X@{}}
    \toprule
    & \textbf{Stage 1} & \textbf{Stage 2} & \textbf{Stage 3} \\
    \midrule
    Rows per table & 1,024 & 400--10,240 & 200--60,000 \\
    Train rows (share) & 30--90\% & 79--81\% & 79--81\% \\
    Features & 2--100 & 2--100 & 2--100 (M/L: fewer above 32K/25K rows) \\
    \midrule
    Realism augmentations & none & +relabeled targets & +heavy-tailed targets \\
    & & +coarsened features & \\
    & & +high-cardinality codes & \\
    & & +argmax classes & \\
    Too-easy tables kept & none & half & half \\
    \midrule
    Learning rate, peak to end & $3 \times 10^{-4}$ to 0 & $4 \times 10^{-5}$ to $4 \times 10^{-8}$ & $1 \times 10^{-5}$ to $1 \times 10^{-7}$ \\
    Gradient-norm clipping & 10 & 10 & 1 \\
    Precision & BF16; FP16 for regressors & BF16 & BF16 \\
    \midrule
    \multicolumn{4}{@{}l}{\emph{Steps $\times$ tables per step}} \\
    \quad \kumoS & 500K $\times$ 64 & 40K $\times$ 128 & 12K $\times$ 128 \\
    \quad \kumoM & 500K $\times$ 128 & 40K $\times$ 128 & 12K $\times$ 128 \\
    \quad \kumoL & 1M $\times$ 128 & 60K $\times$ 128 & 18K $\times$ 128 \\
    \midrule
    \multicolumn{4}{@{}l}{\emph{GPU-hours, classifier / regressor}} \\
    \quad \kumoS & 369 / 361 & 227 / 225 & 196 / 195 \\
    \quad \kumoM & 1,935 / 1,931 & 946 / 969 & 749 / 656 \\
    \quad \kumoL & 7,212 / 7,074 & 2,285 / 2,285 & 2,120 / 2,172 \\
    \bottomrule
  \end{tabularx}
\end{table}

\paragraph{Training Setup.}
Training cost grows with the number of rows, so stages move from many short tables to fewer, longer ones.
The first stage trains on tables of 1,024 rows. It takes over 90\% of the steps but at most 62\% of the GPU-hours.
The second stage draws 400 to 10,240 rows per table, while the third stage draws 200 to 60K rows, with fewer features on its longest tables to bound memory.
In stages 2 and 3 we also lower the learning rate and add the realism augmentations described in \cref{sec:prior}. 
Each stage starts from the final weights of the previous stage. 
The three model sizes differ in the number of steps, the batch size, and the feature cap on the long tables:
\kumoL takes twice as many training steps as the smaller variants in stage~1 and 1.5 times as many in later stages, with fewer features in its longest tables.
Training a single checkpoint took about 800, 3,600, and 11,600 GPU-hours for the three sizes of \modelname (\cref{tab:training}).

\paragraph{Objectives.}
Classifiers minimize the cross-entropy of the test labels over classes present in the context up to at most ten classes.
Regressors minimize the pinball loss~\citep{koenker1978regression}, averaged over 999 quantile levels, on targets standardized using the mean and standard deviation of context targets.

\paragraph{Optimization.}
Most weight matrices are optimized with Muon~\citep{jordan2024muon}, while the remaining parameters are optimized with AdamW~\citep{loshchilov2019decoupled}. We apply decoupled weight decay of 0.01 to most parameters, which pulls the temperature coefficients $\alpha_h$ toward zero.
Classifiers train in BF16 throughout; regressors use FP16 with dynamic loss scaling~\citep{micikevicius2018mixed} in the first stage and BF16 in later stages.
BF16 caused the first ICL layer's mean temperature coefficient to grow to about 30 times its FP16 value, degrading prediction quality in the first stage. In stages 2 and 3, regression checkpoints trained with BF16 had lower RMSE than their FP16 counterparts on most development datasets.

\section{Inference}
\label{sec:inference}
\suppressfloats[t]

\modelname predicts on new tables without further training.
Each forward pass uses the table's training rows to predict its test rows.
The inference stack adds preprocessing, ensembling, and key/value caching:
\vspace{-0.1cm}
\paragraph{Preprocessing.}
We identify each column as numerical or categorical and fit all preprocessing steps on training rows only.
Categorical values are integer-encoded in lexicographic order and treated as numerical values downstream.
As in pretraining, all input features are standardized and clipped.

\paragraph{Estimators.}
Preprocessing affects the computation of a forward pass:
\textbf{(1)} the scale of a feature changes its sine and cosine representation, \textbf{(2)}
the column order sets the rotary positions and feature groups of a token, and \textbf{(3)}
the encoding of the classes sets the label embeddings.
We therefore average $E$ estimators that each make these choices differently, with $E=8$ for \kumoS and \kumoM and $E=16$ for \kumoL.
For feature scaling, the estimators take turns among four monotone maps, namely the identity, the Yeo--Johnson transform~\citep{yeo2000power}, the squashing scaler~\citep{holzmuller2024better}, and a rank-Gaussian map.
Estimators may also flip the signs of numerical features and regression targets at random.
For the column order, each estimator orders the columns by its own row of a Latin square and keeps the first 500.
For the classes, each estimator shifts the class codes cyclically.
Additionally, on datasets with more than 200K rows, each estimator subsamples its training set to 200K rows.

\paragraph{Averaging.}
For classification, \sdm averages the estimators' logits and then applies the softmax. Unlike other models, we set our default softmax temperature to $1$.
With more than ten classes, each estimator first splits the task into ten-class subtasks with error-correcting output codes~\citep{dietterich1994ecoc}.
For regression, each estimator predicts 999 quantiles.
At each quantile level, \sdm drops the highest and lowest 20\% of the estimators and averages the rest.

\paragraph{Cache.}
\label{sec:inference-engine}

Training rows never attend to test rows, so keys and values of each transformer block can be cached for fast batch inference.
With several estimators, we keep the caches in pinned host memory and copy the next estimator's cache to the GPU while the current one runs.

\section{Benchmark Results}
\label{sec:results}
\suppressfloats[t]

We evaluate \modelname on eight tabular benchmarks, covering broad suites of real-world tabular datasets~\citep[TALENT, TabArena, and BeyondArena;][]{ye2024closer,erickson2025tabarena,purucker2026beyondiid}, probabilistic regression~\citep[ScoringBench;][]{landsgesell2026scoringbench}, tables with strings, text, or images~\citep[STRABLE and \mbox{MulTaBench};][]{blayer2026strable,arazi2026multabench}, and wide scientific tables~\citep[RamanBench and \mbox{TabBench-Bio};][]{koddenbrock2026ramanbench,kreuer2026tabbenchbio}.

\begin{figure}[tbp]
  \centering
  \includefig{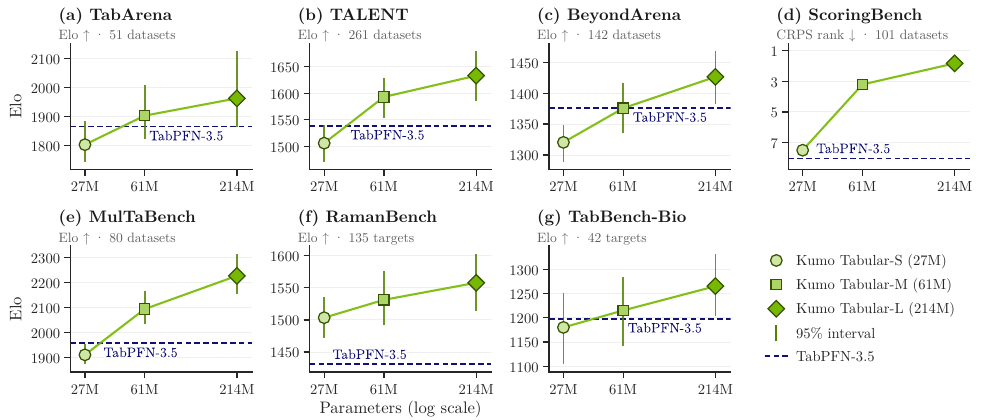}
  \caption{\textbf{Benchmark scores by model size.}
    Elo with 95\% interval, or mean CRPS rank of each model size plotted against its number of model parameters.
    }
  \label{fig:scaling}
\end{figure}

\kumoL ranks first on all benchmarks (\emph{cf.}~\cref{fig:hero}), with predictive performance improving across model sizes on each of the benchmarks (\cref{fig:scaling}).
Its lowest ranks occur on grouped splits and tables with more than 100K training rows, identifying two limitations beyond the pretraining distribution.

\vspace{-0.3cm}
\paragraph{Protocol.}
\modelname runs with the released weights without per-dataset tuning.
Each benchmark determines the data splits and evaluation metrics, summarized with the leaderboard snapshots in \cref{tab:app-benchmarks}.
Unless noted otherwise, we use FP16 autocast with 8 estimators for \kumoS and \kumoM and 16 for \kumoL.\footnote{The MulTaBench and RamanBench leaderboards do not report the estimator counts or numerical precision of their \modelname entries. In our TALENT and \mbox{TabBench-Bio} runs, classes were assigned codes in order of first appearance. The predicted class probabilities therefore depend slightly on the order of the train rows.}
Six leaderboards rank by Elo, a rating fitted to pairwise comparisons between entries on each split or target~\citep{bradley1952rank,elo1978rating}.
ScoringBench ranks by mean rank in the continuous ranked probability score (CRPS), and STRABLE by the median score over table-folds, $R^2$ for regression and ROC AUC for classification.
Each Elo ranking is fitted separately, so ratings are comparable only within that ranking.
TabArena, BeyondArena, and RamanBench replace missing results with RandomForest scores. MulTaBench does not impute missing results. We report median runtimes per 1,000 rows on each benchmark's hardware.
Comparisons on matched hardware are identified explicitly, since entries may use different devices.
For \modelname, fit time includes processing the train rows, and prediction time covers the test rows.

\paragraph{Uncertainty.}
The statistical evidence for a lead varies across benchmarks and tests.
The first-place rankings are point estimates, and on all six Elo leaderboards, the 95\% confidence interval of \kumoL overlaps that of the highest-ranked entry outside the \modelname family.
Against that entry, paired comparisons favor \kumoL on TALENT, ScoringBench, STRABLE, and \mbox{TabBench-Bio} under exact two-sided sign tests (unadjusted $p < 0.05$, \cref{tab:results}).
Against \mbox{TabPFN-3.5}, a strong competitor across all eight benchmarks, \kumoL wins more comparisons than it loses on every benchmark.
All five benchmarks with paired results by dataset, table, or target favor \kumoL under the same exact two-sided sign test (unadjusted $p < 10^{-3}$).
For the remaining three, the reported win rates are 74\% on TabArena, 67\% on BeyondArena, and 71\% on RamanBench.

\suppressfloats[t]
\begin{table}[!ht]
  \captionsetup{position=top}
  \caption{\textbf{The eight benchmarks.} Under each name, the leaderboard version and the date of access in 2026 (TALENT and \mbox{TabBench-Bio}: the date we scored them). Datasets: classification/regression. Rows: per dataset (TALENT and BeyondArena: training rows), with the most a model sees in parentheses where the benchmark subsamples. Columns: features (BeyondArena: after preprocessing). Entries: each configuration of a method counted once.}\label{tab:app-benchmarks}
  \centering\footnotesize
  \setlength{\tabcolsep}{3pt}
  \begin{tabularx}{\linewidth}{@{}>{\raggedright\arraybackslash}p{1.9cm}>{\raggedright\arraybackslash}p{1.35cm}>{\raggedright\arraybackslash}p{1.6cm}>{\raggedright\arraybackslash}p{1.45cm}>{\raggedright\arraybackslash}X>{\raggedright\arraybackslash}p{1.55cm}>{\raggedright\arraybackslash}p{2.65cm}>{\raggedright\arraybackslash}p{1.2cm}@{}}
  \toprule
  Benchmark & Datasets (cls./reg.) & Rows & Columns & Splits & Metrics & Ranking; uncertainty & Entries \\
  \midrule
    \mbox{TabArena}\newline{\scriptsize v0.1.9.8, 19~Oct} & 51 (38/13) & 748--150,000 & 4--1,776 & 3-fold CV, 10 or 3 repeats; 816 splits & ROC AUC, log-loss, RMSE & Elo, RandomForest = 1000; bootstrap & 97 \\
    \mbox{TALENT}\newline{\scriptsize scored 25~Sep} & 261 (172/89) & 323--97,922 & 3--308 & one fixed split & accuracy, RMSE & Elo, RandomForest = 1000; 200 bootstrap rounds & 41 \\
    \mbox{BeyondArena}\newline{\scriptsize v0.1.9.8, 19~Oct} & 142 (98/44) & 104--1,000,350 & 3--22,215 & random (103), temporal (21), grouped (18); 2,000 splits & ROC AUC, log-loss, RMSE & Elo, XGBoost = 1000; bootstrap & 40 \\
    \mbox{ScoringBench}\newline{\scriptsize 9~Oct ledger, 19~Oct} & 101 (0/101) & 240--$10^7$ ($\leq$3,000) & 3--1,024 & 5-fold CV & CRPS (51 metrics) & mean CRPS rank; Friedman and Nemenyi tests & 62 \\
    \mbox{STRABLE}\newline{\scriptsize v0.1.6, 23~Oct} & 108 (32/76) & 550--75,000 & 4--61, with strings & 3-fold CV & ROC AUC, $R^2$ & median score over 324 table-folds; no intervals & 94 pipelines \\
    \mbox{MulTaBench}\newline{\scriptsize v1.9, 19~Oct} & 80 (40/40) & 400--114,000 ($\leq$10,000) & 2--248, with text or images & 3-fold CV, 2 repeats & 1$-$AUC, log-loss, RMSE & Elo, RandomForest = 1000; 1,000 bootstrap rounds & 26 \\
    \mbox{RamanBench}\newline{\scriptsize v1.4, 19~Oct} & 61; 135 targets (21/114) & 12--130,061 ($\leq$10,000) & 114--11,689 & 3-fold CV, group-aware & ROC AUC, log-loss, RMSE & Elo, Random Forest = 1000; 200 bootstrap rounds & 63 \\
    \mbox{TabBench-Bio}\newline{\scriptsize v0.3.0, 25~Oct} & 43 (30/13) & 71--51,859 & 1,142--244,781 & 5 folds; 28 settings of training rows and features & macro-F1, RMSE & Elo, Random Forest = 1000; 2,000 bootstrap rounds & 33 \\
  \bottomrule
  \end{tabularx}
\end{table}

\begin{table}[tb]
  \centering
  \caption{\textbf{Benchmark results.}
  Brackets show 95\% uncertainty intervals where available.
  The strongest other entry reports the highest-ranked model outside the \modelname family.
  Head-to-head reports \kumoL's wins--losses against that model, excluding ties, with exact two-sided sign-test $p$-values in parentheses.
  Comparisons are across datasets, except for STRABLE (tables) and \mbox{TabBench-Bio} (targets). TabArena and BeyondArena instead report win rates across splits, counting ties as half a win.
  Each dataset contributes equally to these win rates. For STRABLE, the reported score is the median over 324 table-folds, using $R^2$ for regression and ROC AUC for classification. Both compared pipelines use the TILE encoder.
  S/M/L gives the leaderboard ranks of \kumoS, \kumoM, and \kumoL variants.
  Dashes indicate unavailable results.
  }
  \label{tab:results}
  \small
  \setlength{\tabcolsep}{3pt}
  \begin{tabular*}{\linewidth}{@{\extracolsep{\fill}}lllllll@{}}
    \toprule
    \multirow{2}{*}{\textbf{Benchmark}} & \multirow{2}{*}{\textbf{Metric}} & \textbf{Kumo} & \multicolumn{2}{l}{\textbf{Strongest other entry}} & \multirow{2}{*}{\textbf{Head-to-head}} & \multirow{2}{*}{\textbf{S/M/L}} \\
    & & \textbf{Tabular-L} & Name & Score & &  \\
    \midrule
    TabArena & Elo ($\uparrow$) & 1962 [1862, 2124] & LimiX-2 & 1952 [1866, 2082] & 49\% win rate & 5\,/\,3\,/\,1 \\
    TALENT & Elo ($\uparrow$)& 1633 [1585, 1679] & LimiX-2 & 1576 [1536, 1614] & 132--95 (0.02) & 6\,/\,2\,/\,1 \\
    BeyondArena & Elo ($\uparrow$) & 1426 [1382, 1468] & TabPFN-3.5 & 1376 [1341, 1416] & 67\% win rate & 5\,/\,3\,/\,1 \\
    ScoringBench & CRPS rank ($\downarrow$) & 1.84 & LimiX-2 & 4.96 & 68--33 ($6{\times}10^{-4}$) & 4\,/\,2\,/\,1 \\
    STRABLE & Median score ($\uparrow$) & 0.8325 & Causilo & 0.8190 & 95--13 ($<10^{-15}$) & --\,/\,--\,/\,1 \\
    MulTaBench & Elo ($\uparrow$) & 2227 [2155, 2316] & LimiX-2 & 2159 [2081, 2249] & 40--40 (1.0) & 7\,/\,3\,/\,1 \\
    RamanBench & Elo ($\uparrow$) & 1558 [1514, 1603] & Chakra-Tab (full) & 1512 [1472, 1557] & -- & 5\,/\,2\,/\,1 \\
    \mbox{TabBench-Bio} & Elo ($\uparrow$) & 1265 [1204, 1332] & TabPFN-3.5 & 1198 [1151, 1245] & 32--9 ($4{\times}10^{-4}$) & 4\,/\,2\,/\,1 \\
    \bottomrule
  \end{tabular*}
\end{table}

\Needspace{6\baselineskip}
\subsection{Broad Suites of Real Tables}
\paragraph{TabArena} contains 51 curated datasets with up to 150K rows.
\kumoL ranks first among the 97 single-model entries and retains first place in rankings that include AutoML systems.
The strongest tuned and ensembled networks and trees rank more than 300 Elo below every \modelname model size.
\kumoL and \mbox{LimiX-2} have a nearly even head-to-head record, while \kumoL predicts about \taLimixPredictRatio$\times$ faster when both are timed on NVIDIA RTX PRO 6000 GPUs.
The protocol fits eight bagged fold models for each \modelname entry and refits a model on all training rows. Prediction uses the refitted model and its cached context.
\vspace{-0.3cm}
\paragraph{TALENT} provides one fixed split per dataset. Our main evaluation covers 261 datasets after following TabICLv2's selection and requiring results from every entry.
We exclude TabICLv2's 26 development datasets, evaluate the 12 datasets with more than ten classes separately, and omit one further dataset on which \mbox{LimiX-2} did not finish.
We fit \modelname on the training split without TALENT's imputation step and compute Elo with TabArena's leaderboard code.
\kumoL and \kumoM rank first and second of 41 entries, ahead of 25 models that are tuned over 100 trials per dataset with results averaged over 15 seeds.
Paired by dataset, the lead over \mbox{LimiX-2} comes from regression, where \kumoL has the lower error on 58 of the 89 datasets.
In classification, the two do not separate by accuracy, but \kumoL has the lower log-loss on 102 of the 172 datasets ($p = 0.018$).

\paragraph{BeyondArena} applies TabArena's protocol to 142 datasets with up to 1M training rows, using random, temporal, or grouped splits. We use the official \textit{core2k} subset, which allocates 2,000 splits across the datasets, with the same bagging and refitting procedure as on TabArena.\footnote{For tables with more than 10K training rows, the fit limit was 16 hours for \modelname and 4 hours for other TFMs. The longest \modelname runs took around 4 hours.}
\kumoL leads \mbox{TabPFN-3.5} and \kumoM by 50 Elo and predicts about \baTabpfnPredictRatio$\times$ faster.

\subsection{Probabilistic Regression}

\paragraph{ScoringBench} scores predictive distributions with proper scoring rules~\citep{gneiting2007strictly} on 101 regression datasets, each subsampled to at most 3,000 rows.\footnote{Our rankings include the 22 fine-tuned variants omitted from the website's default view. Of the 51 recorded metrics, 37 are scoring rules or point metrics and 14 are diagnostics. The benchmark's shared conversion routine maps \modelname's 999 predicted quantiles to predictive distributions.}
By mean CRPS rank, \kumoL and \kumoM take the first two of 62 places, and \kumoL has a lower CRPS than \mbox{LimiX-2} on 68 of the 101 datasets. The magnitude of the difference is often small, with the two models' CRPS values within 1\% on 59 of the 101 datasets.
\kumoL ranks first on 28 of the 37 scoring rules and point metrics, including every interval score and point metric, while \mbox{TabPFN-3.5} and \mbox{LimiX-2} lead the other nine, which score the predicted density itself or approximate the log score. %

\subsection{Tables with Strings, Text, or Images}
\modelname reads text and images through separate encoders.
With off-the-shelf general-purpose encoders, \kumoL leads on string and free-text tables, making this combination a strong default.

\paragraph{STRABLE} ranks 94 pipelines on 108 tables with raw string columns.
With the TILE string encoder~\citep{lefebvre2026tile}, \kumoL ranks first and beats the second pipeline, \mbox{TILE + Causilo}, on 95 of the 108 tables.
Even with cheap term frequency--inverse document frequency (Tf-Idf) features, \kumoL has the highest median $R^2$ on regression tables and ROC AUC on classification tables.
Across the five encoders it ran with, its median $R^2$ varies by less than 0.01, against 0.03 for \mbox{TabPFN-3.5}, which it beats with every encoder on 96 to 102 of the 108 tables.

\paragraph{MulTaBench} contains 80 datasets with free-text or image columns.
For \modelname, these columns are embedded with frozen e5-small text and DINOv3-small image encoders~\citep{wang2022e5,simeoni2025dinov3}, then reduced to 30 principal components per column.
Entries are evaluated without bagging.
\kumoL ranks first, above every model that reads the raw text end to end.
Paired by dataset, it is even with \mbox{LimiX-2}.
By Elo, it ranks first on the datasets with text columns, alone or with images, and second on the 18 whose extra columns are images.

\subsection{Wide Scientific Tables}

\modelname ranks first on tables with thousands of features, even though it was pretrained on at most 100 columns.
\paragraph{RamanBench} contains 135 targets from 61 datasets of Raman spectra (median 270 rows and 1,836 features).
The benchmark evaluates default model configurations through AutoGluon, using three-fold bagging and a 10-minute budget for each fit.
\kumoL and \kumoM rank first and second, ahead of Chakra-Tab, and all three sizes rank above RamanPFN, a model built for Raman spectra and developed on an earlier version of the benchmark that contained 52 of its 61 current datasets.

\paragraph{TabBench-Bio} contains 43 biomedical datasets (median 546 rows and 12,968 features) in 28 settings that fix the number of training rows and cap the number of features a model reads.
We follow the benchmark's feature limits and 48\,GB memory budget, removing \modelname's default cap of 500 columns per estimator.\footnote{The published \kumoM results use an L40S GPU. We evaluated \kumoS and \kumoL with the same harness and adapter on H100 GPUs limited to 48\,GB. Including these entries changes the Elo of each entry covering all targets by at most 3 points.}
We use the strict ranking, in which failed runs receive the score of a constant predictor.
At the benchmark's reference setting (100 training rows, at most 10K features), \kumoL and \kumoM rank first and second, and \kumoL beats \mbox{TabPFN-3.5} on 32 of the 42 targets that this setting scores.
A \modelname model ranks first in each of the 28 settings, with \kumoL leading in 18, \kumoS in nine, and \kumoM in one.
The nine settings led by \kumoS use at least 200 training rows or all available rows. In none of these settings is its paired advantage over \kumoL statistically significant.

\section{Ablation Study}
\label{sec:ablations}

\begin{figure}[htp]
  \centering
  \includefig{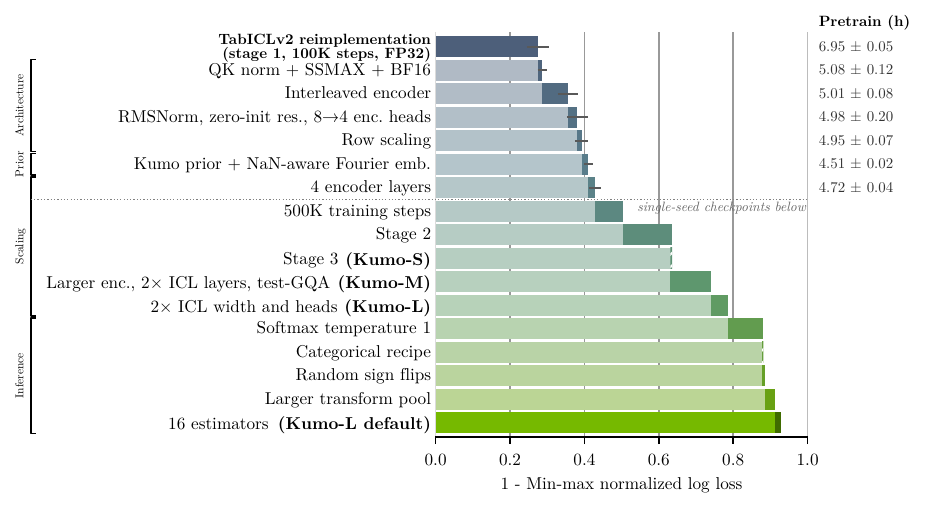}
  \caption{\textbf{Additive ablation from a TabICLv2 reimplementation to \kumoL.}
    Each row builds on the row above it and reports one minus the min-max normalized log loss on 36 TabArena classification datasets with fewer than 1{,}000 features (higher is better).
    On every outer split, the log losses of all models in the figure are rescaled to $[0, 1]$ between the best and the worst, and the result is averaged over splits and datasets.
    The light part of a bar is the score of the row it builds on, and the dark part is the change of the row, hatched if the change is negative.
    The first six rows are runs of 100K steps, with the mean and standard deviation over three pretraining seeds. The remaining rows are single checkpoints from our pretraining runs.
    All models are evaluated on all TabArena outer splits (576 in total).
    The inference rows change only the inference recipe of \kumoL.
    Pretrain (h) is the wall-clock pretraining time on eight RTX PRO 6000 GPUs.}
  \label{fig:ablation-additive}
\end{figure}

We conduct an ablation study on the changes that distinguish \modelname from TabICLv2. 
\cref{fig:ablation-additive} presents the results.
We start from our reimplementation of the TabICLv2 model and prior, trained with our optimization setup (\cref{sec:training}) and a peak learning rate of $3 \times 10^{-4}$.
This setting is the same across the ablations, and we do not expect it to alter our findings.
Because we observed instabilities in longer development runs, we train the TabICLv2 reimplementation in FP32.

\paragraph{Architecture.}
Our first change targets training stability: it adds query-key normalization and replaces QASSMax with SSMax~\citep{nakanishi2025scalable}, applying weight decay to the attention scales.
This makes the runs stable and allows us to perform the rest of the ablations in BF16, which also reduces the training time of the 100K-step runs from 6.95 to 5.08 hours.
Next, we change the layout of the encoder from 3 column-attention layers followed by 3 row-attention layers to our interleaved variant. 
This represents the most significant architectural improvement. 
We attribute this to the model being better equipped to handle interactions between columns.
Next, we slightly adjust the transformer blocks by changing the normalization to RMSNorm and zero-initializing residual connections. 
We also reduce the number of attention heads in the encoder to 4 and double their size to increase training efficiency.
Finally, we introduce bounded QASSMax along the row attention axis.

\paragraph{Prior.}
We next measure the effect of the prior. To account for the missing values added to the data, we also switch to our missing-value-aware embeddings. 
The prior change marginally improves performance. 
Our implementation also speeds up the training across all three stages (from 4.95 to 4.51 hours for the 100K-step runs).
During development, we observed that the prior changes matter more with longer training. Some of them only start in stage 2, which brings the largest improvement in our ablation study.

\begin{figure}[tbp]
  \centering
  \includefig{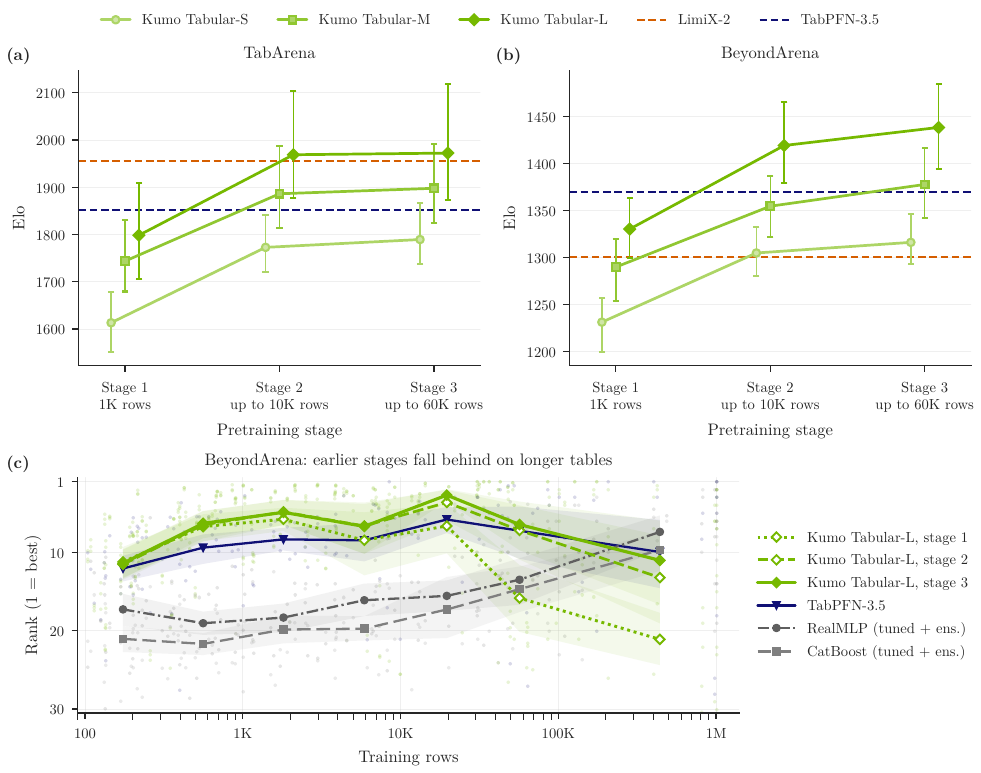}
  \caption{\textbf{Benchmark scores after each pretraining stage.}
    Elo with 95\% intervals on \textbf{(a)} TabArena and \textbf{(b)} BeyondArena.
    \textbf{(c)}~Rank on each BeyondArena dataset by number of training rows.
    Dotted lines report the mean ranks with 95\% intervals shaded.
    }
  \label{fig:curriculum}
\end{figure}

\paragraph{Scaling.}
Next, we look into scaling up the model. 
With the efficiency gains from our architecture and prior changes, we invest more parameters into the encoder, which we identified during development as a key bottleneck in two-stage ICL architectures.
We first increase the layer count from three row and column layers to four, which brings a small improvement at a small cost in training time (from 4.51 to 4.72 hours). 
In our development runs we observe that this is more pronounced when paired with an interleaved architecture.
We then report the full stage-1 and stage-2 checkpoints, which are a significant jump from the 100K-step ablation setup and also carry some of the gains of our prior.
Stage 3 lowers the score in \cref{fig:ablation-additive}. We analyze long-context performance in more depth in \cref{fig:curriculum}.
After stage 3, we arrive at the released \kumoS checkpoint, which we still run with the TabICLv2 inference recipe, with the single modification of not imputing missing values.
We then further scale the encoder and the ICL stack of our model, doubling the channel dimension and number of inducing points, increasing the encoder depth to 6 layers and the ICL depth to 24 layers. 
This represents a significant improvement in performance (winning on 34 of 36 datasets), changing the parameter count of the model from one matching TabICLv2 to one closer to TabPFN-3~\citep{grinsztajn2026tabpfn}, which also includes Test-GQA, allowing for a deeper ICL stack.
Here we also note that \kumoM doubles the batch size of stage 1 relative to \kumoS, and \kumoL doubles both the batch size and the number of steps, to account for the increase in parameter count.
For \kumoM, we also introduce a down-projection layer to keep the width of the ICL stack matching the smaller model. 
We remove this layer for \kumoL, doubling both the number of channels and attention heads.

\paragraph{Inference.}
Up to here, we use the TabICLv2 preprocessing recipe adjusted for missing values. 
Finally, we ablate the inference recipe changes.
For classification with log loss, we observe that changing the softmax temperature to 1 provides the biggest gains.
This matches training, where the cross-entropy loss uses an unscaled softmax. The gains are larger on binary than on multiclass datasets.
We observe that the changes to our categorical recipe are somewhat detrimental to performance on TabArena.
However, we found them useful in practice and helpful across a broader array of datasets.
Randomly flipping the signs of numerical features (and labels for regression) results in a small boost for classification and a more pronounced one for regression in our development runs.
Next, we increase the diversity of numerical preprocessing transforms, which also allows the model to benefit further from more estimators due to the increased diversity of dataset views.
By increasing the number of estimators to 16, we end up with the official \kumoL default checkpoint and recipe.

\paragraph{Long Context Performance.}
We analyze the effects of pretraining on longer artificial input tables.
On TabArena and BeyondArena, stage~2 brings most of the gain in Elo: at every size, stage~2 adds 65 to 170 points, and stage~3 adds at most 23 (\emph{cf.}~\cref{fig:curriculum}).
On BeyondArena, each stage improves \kumoL mainly on datasets with more training rows than the tables of the stage before: its stage-1 checkpoint keeps pace with the released model up to about 1K training rows, and its stage-2 checkpoint up to about 10K.

\section{Accuracy--Efficiency Performance}

On a table with $n$ training rows and $m$ columns, the runtime complexity for fitting \modelname is $O(n^2 + nm^2)$, while prediction with KV-caching takes $O(n + m^2)$ per test point. Hence, as long as $m \ll \sqrt{n}$, the number of columns $m$ does not matter much w.r.t.\ runtime.

\begin{figure}[tb]
  \centering
  \includefig{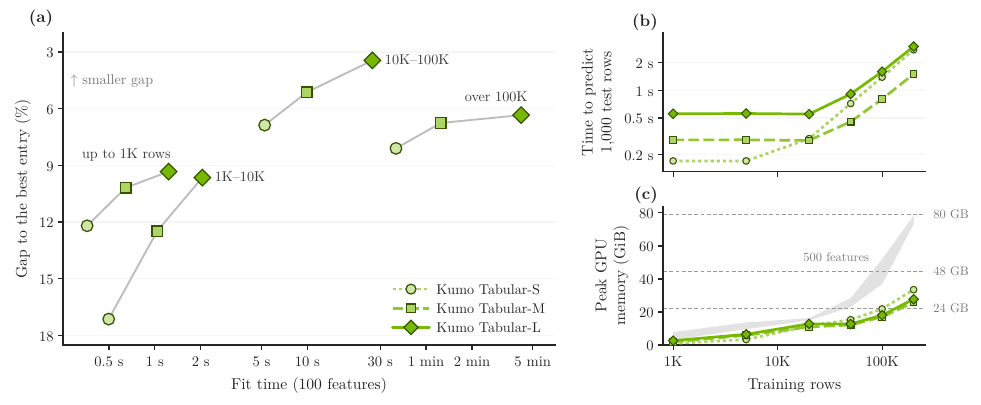}
  \caption{\textbf{Accuracy, runtime, and memory across model sizes.}
    (a)~Average gap to the best entry on the BeyondArena datasets in four ranges of training rows, against the fit time of a table with 1,000, 5,000, 50K, or 200K training rows, respectively.
    For each dataset, the gap is $100(e-e_{\mathrm{best}})/e$, where $e$ and $e_{\mathrm{best}}$ are the errors of the evaluated model and the best entry. We report the mean gap over datasets.
    (b)~Time to predict 1,000 test rows on one H100 with 80 GB VRAM and (c)~peak GPU memory over fit and prediction with 100 features, including the model weights. The shaded area spans the memory for the three sizes with 500 features.
    Costs: binary classification with 100 features, recommended estimators, and FP16 autocast on one H100 80GB.}
  \label{fig:usage}
\end{figure}

\cref{fig:usage} shows the accuracy, runtime, and memory usage of \modelname across different dataset sizes under FP16 autocast with at most 200K train rows and 500 columns per estimator.
We measure the costs on one NVIDIA H100 80GB, on tables with random numerical features and 1,000 test rows.
Regression differs from binary classification by at most 1.5\,GiB of memory and 10\% of time.
We find that the improvement of \kumoL over \kumoM is most pronounced on tables between 1K and 100K rows.
Larger context does not yet show a clear lift, potentially due to the global pretraining limit of 60K rows.

During fitting of the model, the main memory usage typically comes from the table encoder, which holds one token per train cell in memory at once.
During prediction, the GPU also holds the KV-caches of the running and the next estimator (inducing vectors in the encoder, regular KV cache in the ICL layers).
With \kumoL and 100 features, a 24\,GB GPU holds every size up to 100K training rows.
With 500 features, a 24\,GB GPU can hold 20K training rows, a 48\,GB GPU can hold 50K training rows, and an 80\,GB GPU can hold 200K training rows.
At 200K training rows and 100 features, \kumoS, \kumoM, and \kumoL take 48, 26, and 51\,GiB of host memory, respectively.

Prediction stays cheap: with up to ten classes, 1,000 test rows take under 4\,s at every measured table shape.
For tasks with more than ten classes, an estimator runs its ten-class subtasks together as one batch of relabeled copies of the table.

\section{Limitations}

Generalization beyond the pretraining distribution remains uneven.
Pretraining uses i.i.d.\ tables with at most 60K rows and therefore does not cover grouped or temporal dependencies, or longer contexts.
On BeyondArena's grouped splits, \kumoL ranks fourth, behind tuned and ensembled RealMLP, TabM, and LightGBM.
On tables with more than 100K training rows, it ranks ninth, behind seven tuned or tuned and ensembled networks and tree ensembles, together with \mbox{TabPFN-3.5}.
The gap on temporal splits is smaller, with \kumoL ranking third and trailing the leader, \mbox{TabPFN-3.5}, by 3 Elo.
The strongest limitations in this evaluation thus concern generalization to unseen groups and very large contexts.

Large tables also come with large runtime and memory requirements, especially when subsampling is disabled.
Additionally, \modelname can only process single tables, and handles semantic information like text and images through separate encoders. For distributional regression, \modelname excels in quantile-based metrics like CRPS but trails density-native methods on ScoringBench rules that score the predicted density itself.
We used subsets of the TabArena, BeyondArena, and TALENT benchmarks for validation during development of \modelname, but as \modelname also tops five other benchmarks, benchmark overfitting does not seem to be a problem.

\section{Conclusion}
\label{sec:conclusion}

\modelname combines strong predictive performance with efficient inference in a fully open family of tabular foundation models pretrained entirely on synthetic data.
\kumoL ranks first on the eight evaluated benchmarks under their primary metrics, and all three model sizes lie on the Pareto frontiers of accuracy and prediction time on TabArena and BeyondArena.
The remaining gaps on grouped splits and very large tables motivate further study of generalization beyond the pretraining distribution.
By releasing the model weights, synthetic prior, and pretraining and inference code, we provide the community with an experimental foundation for developing more accurate, efficient, and broadly applicable tabular models.

\section*{Acknowledgements}
We thank the maintainers of TabArena, BeyondArena, ScoringBench, STRABLE, MulTaBench, RamanBench, and \mbox{TabBench-Bio} for including \modelname in their benchmarks and publishing the results.
We thank Sascha Marton for help with writing.
David Holzmüller thanks Ingo Steinwart for providing computational resources funded by Deutsche Forschungsgemeinschaft (DFG, German Research Foundation) under Germany’s Excellence Strategy -- EXC 2075 -- 390740016, acknowledges HPC resources from GENCI–IDRIS (Grant 2024 AD011014864R2), and thanks Probabl for providing Claude API access.

{\small
\bibliographystyle{plainnat}
\bibliography{references}
}

\clearpage
\appendix
\counterwithin{figure}{section}
\counterwithin{table}{section}
\startcontents[app]
\section*{Appendix Contents}
{\hypersetup{linkcolor=black}\printcontents[app]{}{1}{\setcounter{tocdepth}{2}}}
\FloatBarrier
\section{The Kumo Prior}
\label{app:prior}
\suppressfloats[t]

\newcommand{\priornew}{\,{\color{kumogreen}\small\textsc{new}}}
\newcommand{\priorchanged}{\,{\color{kumogreen}\small\textsc{changed}}}
\newcommand{\specgroup}[2]{\multicolumn{#1}{@{}>{\raggedright\arraybackslash}p{0.97\linewidth}@{}}{#2}\\*}
\newcommand{\specitem}{\hspace*{0.9em}}

This appendix gives the sampling rules, equations, and stage settings for the prior introduced in \cref{sec:prior}.

\begin{table}[!ht]
  \centering
    \caption{\textbf{Differences between the \priorname and the TabICLv2 prior.} Step refers to the generation steps in \cref{fig:prior}.}
  \label{tab:prior-overview}
  \small
  \begin{tabularx}{\linewidth}{@{}>{\raggedright\arraybackslash}p{0.3\linewidth}>{\raggedright\arraybackslash}Xcc@{}}
    \toprule
    Mechanism & Difference to TabICLv2 & Step & Section \\
    \midrule
    \multicolumn{4}{@{}l}{\emph{Every stage}} \\
    \quad Distributional numerical converter & row-dependent noise, hidden subgroups, bounded values, and counts & 4 & \ref{app:prior-converters} \\
    \quad Correlated feature blocks & groups of columns that share the same noise & 5 & \ref{app:prior-blocks} \\
        \quad Missing values & missing rows, feature-dependent missingness, values outside selected ranges, and random missing cells; stages 2 and 3 also add nearly empty columns and blocks controlled by a latent factor per row & 5 & \ref{app:prior-missing} \\
    \quad Trivial examples & a test for nearly deterministic tables; rejected in stage~1, half kept in stages 2 and 3 & 6 & \ref{app:prior-filter} \\
    \addlinespace
    \multicolumn{4}{@{}l}{\emph{Realism augmentations, stages 2 and 3}} \\
    \quad Argmax classes & categorical columns and class labels that are a deterministic function of their node & 4 & \ref{app:prior-realism} \\
    \quad Relabeled targets & targets recomputed from a few features & 5 & \ref{app:prior-realism} \\
    \quad Coarsened features & features with few levels, so identical rows can carry different labels & 5 & \ref{app:prior-realism} \\
    \quad High-cardinality codes & categorical codes with hundreds of levels of power-law sizes & 5 & \ref{app:prior-realism} \\
    \quad Heavy-tailed targets & skewed regression targets with extreme values, in stage~3 only & 5 & \ref{app:prior-realism} \\
    \addlinespace
    \multicolumn{4}{@{}l}{\emph{Changed parts of TabICLv2, every stage}} \\
    \quad Per-SCM preferences & per-table probabilities for discrete choices, such as the function family of each node & 1 & \ref{app:prior-base} \\
    \quad Unconnected targets & graphs in which no feature shares an ancestor with the target are kept, and the learnability filter rejects most of their tables & 2 & \ref{app:prior-base} \\
    \quad Clusters and products & cluster distances in the $L_1$ or $L_2$ norm, with widths of random spread; products of any two other functions & 3 & \ref{app:prior-base} \\
    \quad Kumaraswamy warp & a warp of the column itself, instead of the node channel that the children read & 4 & \ref{app:prior-base} \\
    \bottomrule
  \end{tabularx}
\end{table}

\subsection{Changes to the TabICLv2 Prior}
\label{app:prior-base}

\Cref{tab:prior-overview} summarizes the additions and changes relative to TabICLv2~\citep{qu2026tabiclv2}.
The changes to the base generator affect hyperparameter sampling, graph selection, node functions, and numerical readout (steps~1--4 in \cref{fig:prior}).

Per-SCM correlated sampling replaces the independent sampling of categorical choices used during TabICLv2 pretraining~\citep[Appendix~E.2]{qu2026tabiclv2}.
These choices include function families, activations, and converters.
A separate probability vector is sampled from a Dirichlet distribution for each choice type and reused for all selections of that type within the SCM.
The shared probabilities can favor repeated use of a few options across nodes or columns, or produce a broader mix.
New probability vectors are sampled for each table.

Unlike TabICLv2~\citep[Section~5]{qu2026tabiclv2}, graphs with no shared ancestor between any feature and the target are retained during graph sampling.
Their generated tables are passed to the learnability filter (\cref{app:prior-filter}), so rejection is based on predictive performance rather than graph structure.

Two node-function families are modified relative to TabICLv2~\citep[Appendix~E.8]{qu2026tabiclv2}.
Cluster functions use either $L_1$ or $L_2$ distances, which are fast to compute, whereas TabICLv2 samples $p$ between 1 and 4 for its $L_p$ distance.
The spread of log cluster widths is sampled instead of fixed at 0.1.
Product functions apply two sampled functions to the same input and multiply their outputs elementwise.
The two functions can be selected from any of the seven non-product families, including MLPs and cluster functions, which TabICLv2 excludes from this selection.

The optional Kumaraswamy warp is applied to the observed numerical column in the \priorname, leaving the hidden channels passed to child nodes unchanged.
In TabICLv2, the hidden channel is warped while the observed column retains its unwarped value~\citep[Appendix~E.6]{qu2026tabiclv2}.

\subsection{Numerical Columns as Noisy Measurements}
\label{app:prior-converters}

\begin{figure}[t]
  \centering
  \includefig{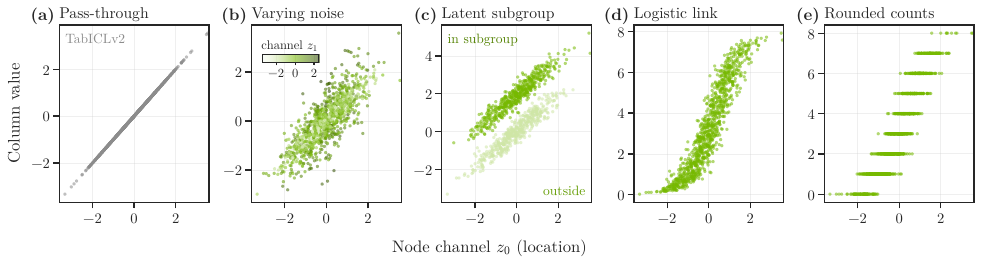}
  \caption{\textbf{Numerical columns read from one node's channels.}
    Each panel converts the same 1,000 rows of three standard normal channels.
    (a)~The pass-through converter of TabICLv2.
    (b--e)~Individual components of the distributional converter: noise that grows with $z_1$ ($w=0.7$, $\gamma=0.5$), a hidden subgroup ($w=0.9$, $\delta=2$), the logistic link ($w=0.9$, $a=1.5$, $b=-0.5$, $c=8$), and the link with rounding.}
  \label{fig:prior-numconv}
\end{figure}

TabICLv2 introduced converters as mechanisms to extract columns of different data types from graph nodes. While its categorical converter contains different sub-mechanisms to extract categoricals, its numerical converter simply reads out the values of a single dimension on the node (and optionally warps the value on the node afterwards), see \cref{fig:prior-numconv}a. With the \emph{distributional numerical converter}, we diversify this process by introducing noise that is controlled by multiple dimensions of a graph node. The goal is to create a variety of typical noise distributions. While TabPFNv1 used Gaussian noise in the causal graph \citep{hollmann2023tabpfn}, foundation models for causal inference and survival analysis have already used larger families of noise distributions \citep[e.g.,][]{balazadeh2025causalpfn, robertson2025pfn, ma2026foundation, seletkov2026survival,qi2026survivalpfn,bohm2026survpfn,alberge2026survicl}. We bring a flexible family of noise distributions to tabular foundation models for classification and regression.

The distributional converter is parameterized by three channels $z_0$, $z_1$, and $z_2$ for location, noise scale, and subgroup probability (\cref{fig:prior-numconv}b--e).
These channels are standardized and clipped with $z \mapsto 5\tanh(z/5)$.
Subgroup membership $g$ and standard normal noise $\varepsilon$ are sampled for each row.
The latent value $\eta$ and observed value $v$ are computed as
\begin{equation}
  \begin{gathered}
    g\sim\mathrm{Bernoulli}\big(\sigma(z_2)\big),\qquad
    \eta=\sqrt{w}\,(z_0+\delta g)+\sqrt{1-w}\;e^{\gamma z_1-\gamma^2/2}\,\varepsilon,\quad \varepsilon\sim\mathcal{N}(0,1),\\
    v=c\,\eta\ \text{(identity link)}\qquad\text{or}\qquad v=c\,\sigma(a\eta+b)\ \text{(logistic link)},
  \end{gathered}
  \label{eq:prior-dist}
\end{equation}
where $\sigma$ is the logistic function.

The signal fraction $w$ controls the balance between signal and noise.
The parameter $\gamma$ controls how noise scales with $z_1$, and $\delta$ shifts the subgroup's location.
The scale $c$ controls the output magnitude.
The logistic link has slope parameter $a$, offset $b$, and output range $(0,c)$.
The output is rounded to an integer with probability 0.3.
A small $c$ with the logistic link produces counts with few levels.

The distributional converter is selected for a column when a correlated draw is below 0.1, so with roughly 10\% probability.
The number of columns using this converter can range from zero to all numerical columns in a table.
Optional feedback is implemented by writing the sampled value back into the node channels used by its children.

\subsection{Correlated Feature Blocks}
\label{app:prior-blocks}

\begin{figure}[t]
  \centering
  \includefig{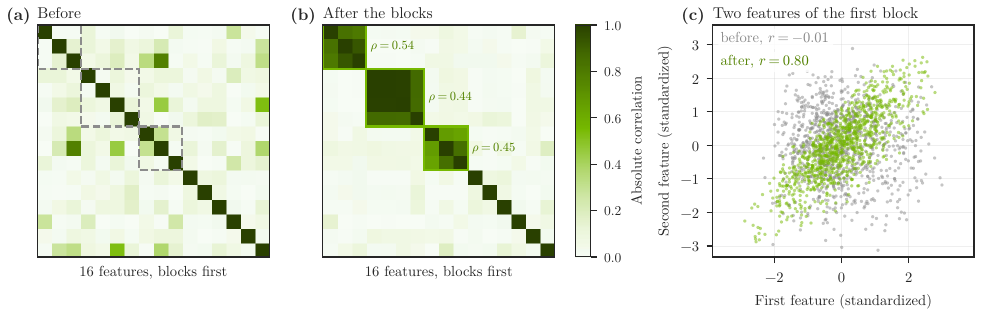}
  \caption{\textbf{Correlated feature blocks.}
    Absolute correlations between the 16 numerical features of one drawn table (a)~before and (b)~after three blocks with strengths $\rho$ of 0.54, 0.44, and 0.45; (c)~two features of the first block.}
  \label{fig:prior-blocks}
\end{figure}

Correlated feature blocks are introduced in 30\% of tables (\cref{fig:prior-blocks}).
Between 2 and $m$ columns are selected and divided into 1 to 4 blocks, where $m$ is the number of features.
A strength $\rho\in(0.2,0.7)$ and one shared factor $z$ per row are sampled for each block containing at least two columns.
Numerical values are transformed as $x\mapsto(1-\rho)\,x+\rho\,z$~\citep{synthefy2026nori}.
Categorical labels are reordered to follow $z$, preserving the count of each category.
The transformation is applied before standardization.
The shared factor has a stronger effect on columns with low variance, which can closely follow $z$ after standardization (\cref{fig:prior-blocks}b, $\rho=0.44$).
Its effect is weaker for columns with high variance.

\subsection{Realism Augmentations}
\label{app:prior-realism}

\begin{figure}[t]
  \centering
  \includefig{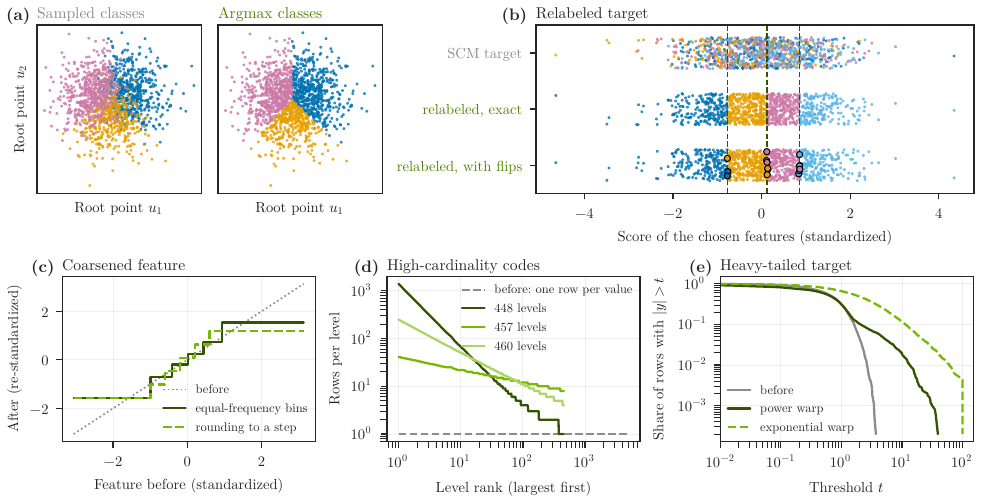}
  \caption{\textbf{Realism augmentations of stages 2 and 3.} Each panel forces one augmentation on; gray: before, green: after.
    (a)~Sampled and argmax classes of 1,500 rows from the same softmax logits, at their two-dimensional root points $(u_1,u_2)$; 82\% agree.
    (b)~Classes before and after relabeling, against the score of the chosen features; circled: flipped labels.
    (c)~One column coarsened to six levels in each mode.
    (d)~Level sizes of three recoded columns.
    (e)~Tails of a regression target under each warp.}
  \label{fig:prior-realism}
\end{figure}

Realism augmentations are enabled from stage~2, except for the heavy-tail transformation, which is enabled in stage~3 (\cref{fig:prior-realism}).
\begin{itemize}[leftmargin=1.2em,itemsep=2pt,topsep=3pt]
  \item \textbf{Argmax classes.}
  Argmax selection is used with probability 0.15 in softmax converters for categorical features and classification targets (\cref{fig:prior-realism}a).
  The category with the highest score is selected instead of sampling from the softmax probabilities.
  \item \textbf{Relabeled targets.}
  Targets are recomputed from 1 to 4 features in 10\% of tables using a sampled linear map, MLP, or Gaussian process (\cref{fig:prior-realism}b).
  Rows are assigned to classes by thresholding their scores, with thresholds chosen to produce randomly sized classes.
  Optional noise in the score ranks can change labels near class boundaries.
  \item \textbf{Coarsened features.}
  A random share of columns is reduced to 2 to 8 levels in 10\% of tables using equal-frequency bins or rounding to fixed steps (\cref{fig:prior-realism}c).
  \item \textbf{High-cardinality codes.}
  Rank-based recoding is applied to 1 to 3 columns in 5\% of tables with at least 201 rows (\cref{fig:prior-realism}d).
  Each column is split into more than 200 categories with power-law category sizes and marked as categorical.
  \item \textbf{Heavy-tailed targets.}
  A power or exponential transformation is applied to the standardized target in 20\% of regression tables in stage~3 (\cref{fig:prior-realism}e).
\end{itemize}

\subsection{Missing Value Patterns}
\label{app:prior-missing}

\begin{figure}[tp]
  \centering
  \includefig{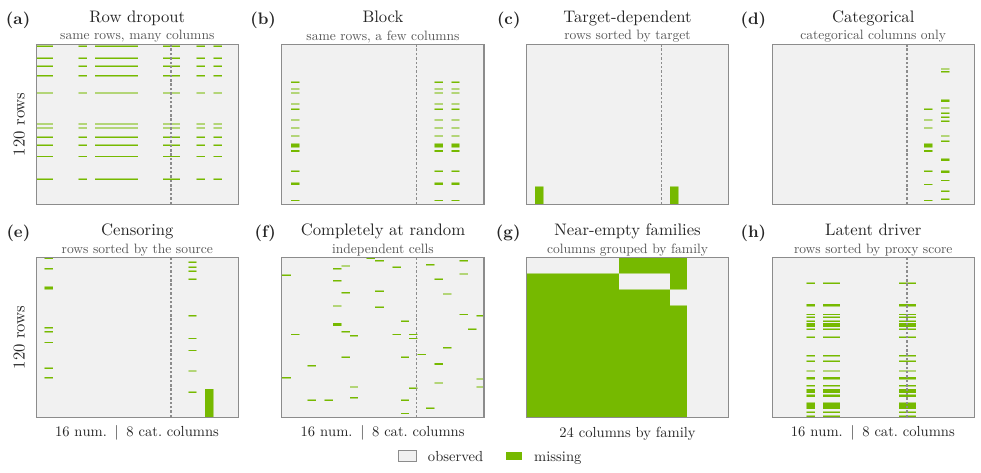}
  \caption{\textbf{Missing-value patterns of the \priorname.}
    Each panel applies one pattern, with the prior's settings, to the same drawn regression table of 120 rows and 24 features, 16 numerical and then 8 categorical (dashed line).
    In~(e), rows are sorted by the numerical column whose tail censors a categorical column; the other censored columns follow other sources, so their missing cells look scattered.}
  \label{fig:prior-missing}
\end{figure}

Missing values are introduced after target generation and before the split into train and test sets.
Each selected mechanism samples the amount of missingness and the affected rows or columns.
Near-empty missingness is selected first and creates columns with high missingness (\cref{fig:prior-missing}g).
If selected, structured missingness (\cref{fig:prior-missing}a--e) is skipped for the entire table.
Random and latent-factor masking are selected independently (\cref{fig:prior-missing}f,h).
Structured and random missingness are used in every stage, while near-empty and latent-factor masking are added in stages~2 and~3.
At least one missingness mechanism is selected for about 21\% of generated tables in stage~1 and 27\% in stages~2 and~3. The target is not masked.
Missing feature values are passed to the network as NaN.

\begin{itemize}[leftmargin=1.2em,itemsep=2pt,topsep=3pt]
  \item \textbf{Structured missingness} (probability 0.2, in tables without near-empty columns).
  One of four masks is applied with probability 1/2.
  The same rows are masked in most columns under \emph{row dropout} (\cref{fig:prior-missing}a) or in a few columns under \emph{block} masking (\cref{fig:prior-missing}b).
  A few columns are masked in rows from one target tail or one class under \emph{target-dependent} masking (\cref{fig:prior-missing}c).
  Values are removed from a few categorical columns under \emph{categorical} masking (\cref{fig:prior-missing}d).
  Subsequent \emph{censoring} is based on the upper or lower tail of a column's own values or another column's values (\cref{fig:prior-missing}e).
  Censoring is applied to a few columns with probability 1 if no cells are missing yet, or with probability 1/2 otherwise.
  \item \textbf{Completely at random} (probability 0.01).
  A probability between 1\% and 8\% is sampled once per table.
  Each feature cell is masked independently with this probability (\cref{fig:prior-missing}f).
  \item \textbf{Near-empty column families} (probability 0.05).
  Observations are restricted to 1--10\% of rows for 50--80\% of columns.
  The same observed rows are used for columns in the same family (\cref{fig:prior-missing}g).
  \item \textbf{Latent driver} (probability 0.03).
  A score is computed for each row as a randomly weighted sum of a few columns, then standardized and perturbed with Gaussian noise.
  Rows with the highest scores form a group in which selected other columns are more likely to be missing (\cref{fig:prior-missing}h).
\end{itemize}

\subsection{Data Quality Checks and the Learnability Filter}
\label{app:prior-filter}

Table generation combines data quality checks with a learnability filter to select tables for training.
First, constant columns are removed.
Classification tables are then checked for matching classes across the train and test sets, while regression tables are checked for sufficient target variation.
Tables that pass these checks are also evaluated by the learnability filter during the first five generation attempts.
If generation fails or a table is rejected, another table is drawn with the same sampled shape.
Generation continues until a table is accepted or 100 attempts have been made.

The learnability filter uses a predictive model consisting of 25 extremely randomized trees~\citep{geurts2006extremely} with a maximum depth of 6, following TabICLv2~\citep[Appendix~E.14]{qu2026tabiclv2}.
Out-of-bag error is compared with the error of predicting the mean, using one-hot labels for classification.
Tables on which the trees do not reliably beat this baseline are rejected in every stage.
Tables with at most 5\% of the baseline error are considered \emph{too easy}.
These tables are rejected in stage~1; half are retained in stages~2 and~3 to include some nearly deterministic tasks.
The fraction of tables classified as unlearnable is 27--37\% for tables with 1,024 rows.
The too-easy rate is 7--8\% for classification and 12\% for regression.

\FloatBarrier
\section{Training Stability}
\label{app:stability}

During development we encountered excessive growth in the attention scales, the learned factors that multiply the attention logits, in longer runs and larger models.
We added query--key normalization (QK-norm), which helped stabilize training by removing the direct dependence of logit scale on query and key projection magnitudes.
However, QK-norm with learned affine gains still allowed the unbounded attention scales to grow.
As the query and key gains multiply with the scales, their growth could compound even after normalization.

\begin{figure}[!htb]
  \centering
  \includefig{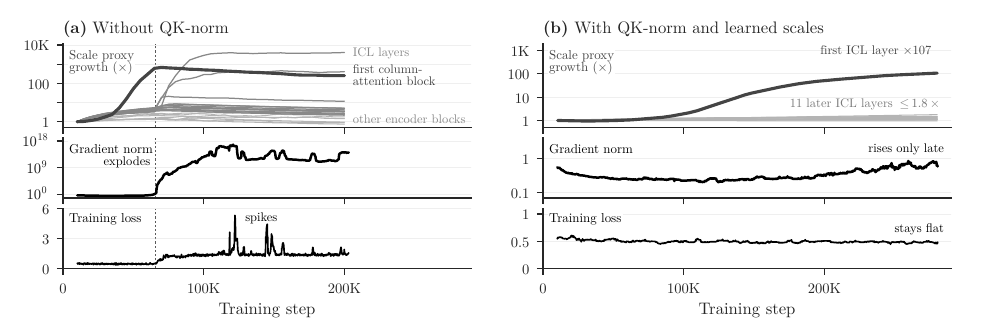}
  \caption{\textbf{Two signatures of training instability.}
    Two BF16 classification runs from development, without QK-norm (a) and with QK-norm and additional learned scales (b).
    Top: parameter-based scale proxies relative to step 10K; middle: pre-clip gradient norm; bottom: training loss.
    \textbf{(a)}~Without QK-norm, the proxy is the product of the largest singular values of the query and key projections.
    \textbf{(b)}~With QK-norm, it is constructed from the RMS values of the learned query and key gains and per-channel query-scale parameters. %
    Gradient norms and losses are shown as rolling medians.}
  \label{fig:divergence}
\end{figure}

\Cref{fig:divergence} illustrates the behavior we encountered.
Without QK-norm, scale growth was followed by abrupt increases in gradient norm and training loss.
With QK-norm, the learned gains and separate per-channel query scale continued growing in the first ICL layer, while the training loss remained nearly constant and the gradient norm rose only later.
Monitoring these parameters and gradient norms exposed changes that were not apparent from the loss alone.

\paragraph{Attention Parameterization.}
\label{app:abl-stability}
These observations led us to remove the learned QK-norm gains and the separate per-channel query scale.
In the released model, QK-norm has no learned gains, and length-dependent scaling has one learned coefficient $\alpha_h$ per head.
We apply weight decay to this coefficient to discourage unchecked growth.
Heads can still learn different attention scales, and the length dependence helps maintain concentrated attention as contexts grow.
This combination was stable in our training runs and enabled us to train with mixed precision using FP16 and BF16.

\paragraph{Possible Feedback Mechanism.}
\label{app:stability-loop}
Some ICL heads may benefit from concentrating attention on a few relevant training rows, providing a reason for attention scales to grow even in stable runs.
We suspected that this concentration could also lead to a feedback loop through attention output magnitude.
When diffuse attention averages value vectors that partially cancel, concentrating the weights can increase the output norm.
In a pre-norm block, that output is added to the residual stream without further normalization.
If the larger contribution reduces the loss, optimization may favor still larger attention scales, reinforcing the concentration and output growth.
We hypothesized that cutting this feedback loop would restore stability. Our parameterization does so simply, without sacrificing performance.

We also tested normalizing the attention output before the residual addition to limit its magnitude independently of attention concentration.
This stabilized development runs but lowered accuracy in short ablations, so we did not retain it in the released model.

\end{document}